\documentclass[oneside]{article}
\usepackage{epsfig}
\usepackage{amssymb,color}
\usepackage{framed}
\usepackage{epsf}
\usepackage{algorithm, algorithmic}
\usepackage{booktabs}
\usepackage{multirow}
\usepackage{graphics,epsfig}
\usepackage[normalem]{ulem}
\usepackage{url}

\newcommand{\qed}{\mbox{}\hspace*{\fill}\nolinebreak\mbox{$\rule{0.6em}{0.6em}$}
}

\definecolor{gray}{rgb}{0.5,0.5,0.5}

\begin{document}
\title{The Mind Grows Circuits}
\author{
Rina Panigrahy, Li Zhang \\
Microsoft Research, Mountain View, CA \\
\{rina, lzha\}@microsoft.com
}
\maketitle
\newcommand{\Omg}{$\Omega$}
\newcommand{\der}{\dot}

\begin{abstract}
There is a vast supply of prior art that study models for mental processes. Some studies in psychology and philosophy approach it from an inner perspective in terms of experiences and percepts. Others such as neurobiology or connectionist-machines approach it externally by viewing the mind as complex circuit of neurons where each neuron is a primitive binary circuit. In this paper, we also model the mind as a place where a circuit grows, starting as a collection of primitive components at birth and then builds up incrementally in a bottom up fashion. A new node is formed by a simple composition of prior nodes when we undergo a repeated experience that can be described by that composition. Unlike neural networks, however, these circuits take ``concepts" or ``percepts" as inputs and outputs. Thus the growing circuits can be likened to a growing collection of lambda expressions that are built on top of one another in an attempt to compress the sensory input as a heuristic to get a gross approximation of its Kolmogorov Complexity.
\end{abstract}

\section{Introduction}
Understanding mental phenomena has been an age old subject in various disciplines ranging from philosophy~\cite{Hume, Locke}, psychology~\cite{emotions}, ethology~\cite{Konrad} cognitive sciences~\cite{Laird} and linguistics~\cite{Torneke}. It is also the subject matter of modern fields such as neuroscience~\cite{neuro} and artificial intelligence~\cite{Minsky, Soar}. Some of the early studies in psychology and philosophy approach the question looking at mental phenomena subjectively in terms of experiences and percepts. Others such as neurobiology or AI approach it from the other end in viewing the mind as a neural network~\cite{neural} where each neuron is a primitive binary circuit. Simplicity theory~\cite{Chater99, CV03} (and other older works ~\cite{Mach}) emphasizes the principle of `simplicity' in cognition -- using the notion of Kolmogorov Complexity~\cite{Kol}, it demonstrates that several cognitive functions can be understood as giving simple descriptions of a certain sensory stimulus. In this paper, we also model the mind as a place where a circuit grows, starting as a collection of primitive components at birth and then grows incrementally bottom up. A new node is formed by a simple composition of prior nodes when we undergo a repeated experience that can be described by that composition. Unlike neural networks, however, these circuits take ``concepts" or ``percepts" as inputs and outputs. Thus the growing circuits can be likened to a growing collection of lambda expressions~\cite{lambda} that are built on top of one another.

\subsection{ Sensations, Emotions and Experience}
Human behavior is rooted in experience. An experience is any sequence of observations, feelings, insights that we go through. Getting a precise understanding of the nature of experiences is key to modeling psychological behavior. At the simplest level it may be a sight, sound or any stimulus from the senses often accompanied by some feeling or emotion. Sensations produce feelings; feelings are diverse but fit into two broad categories -- painful and pleasurable. Intense feelings are the emotions. Sensations -- sights, sounds, smells -- go through a mental evaluation to produce feelings. Sometimes they give rise to `mental concepts'. A sight (say of a scene containing a mountain) may give rise to a concept of a mountain. The feeling that arises from such a sight may be triggered from the concept rather than the sight itself. Concepts and their associated feelings may be stored in memory. Memory may also retain concepts with associated (related) concepts. For example the concept of a mountain may be associated with recent experiences (say hiking trips) related to mountains. External sensations are not necessary to trigger emotions. Concepts retrieved from memory can also trigger emotions. Thoughts is a sequence of concepts retrieved by following a chain of associations. Thoughts may be triggered by an external sensation or simply from recalling concepts from memory -- thoughts produce intense emotions. 

Sensations, feelings, emotions, concepts, thoughts -- these are the objects of the mind. The pleasure principle is an important guiding factor that affects what the mind will be pay attention to. The mind will pay attention to and seek objects that produce pleasurable feelings; it will also pay attention to and push away from objects that produce painful feelings. Another important factor that attracts mental attention is `contrast' -- a change in stimulus -- contrast happens when two very different sensations, feelings or experiences are placed side by side (in time or space). Even a pleasurable feeling from a stimulus tends to fade if the source remains constant. The mind will quickly grow tired and look for alternate new sources of pleasure -- and then grow tired of that again and keep hunting and switching (it may even grow tired of constant hunting and experiencing of pleasure and prefer tranquility for a period). This hunting may occur in external sensations or concepts present in thoughts from memory.
Memory is accumulated over time. But feelings arise from sensations even without memory. A child without memory will also react with feelings on external sights and sounds. Thus the brain (even without any stored memory) maps sights and sounds to feelings. This `circuit' is a primitive circuit that is inborn. For example the taste of milk produces a pleasurable feeling and the taste of grapefruit produces a bitter painful feeling.

This primitive circuit does not have state and produces the same output on the same input, Memory provides state. Over time concepts and associated feelings accumulate in the memory over time and interact with the operation of a circuit. 
The operation of the primitive circuit (without memory) is complex. It is not additive -- two objects together may produce a very different feeling from the simple combination of the feelings produced by each individual object.
Contrast is an important principle in the formation of concepts. The mind creates a concept from anything it sees as different and sees repeatedly. For instance seeing many black dots on a white board will make the mind notice the black dots (due to contrast) and then creates the concept of a 'dot' -- every concept is nothing but the common experience separated by the contrast. Similarly experiencing a very different feeling in time will separate the new feeling based on contrast and create a new concept of that feeling -- which is the experience of that feeling.

An external sensation (such as a sight) produces an experience consisting of several experiences separated by contrast -- these separated experiences create concepts -- repeated similar experiences causes the associated concept to be stored in memory. On a future experience the same or similar experience (and the associated tags) are retrieved from memory.
This is also true when we hear sounds in a language. Syllables are separated by contrasts. A sequence of syllables is an experience which on repetition forms a concept stored in memory. It is associated with the related concept which is the thought of the object that the word semantically refers to. Thus hearing speech can be understood as a collection of experiences separated by contrast and then retrieving related experiences from memory stored as concepts. 

Basic concepts such as time comes from the feeling of `waiting' for something pleasurable; the concept of space comes from `having to reach out' to grab something pleasurable; from the consistent experience that a certain action is always followed by a pleasurable experience -- `if action then reward', comes reason and mental computation.

An important question here is how is a sensation (say a scene) stored in memory~\cite{Smith}. It is unlikely that an image is stored as a raw bit-map. Contrast is used to separate the scene into several scenes. These are in-turn divided recursively using contrast. Some may be mapped to a previously stored concept. Thus the scene is stored recursively as a collection of concepts that can be used to approximately reconstruct the scene. 

\section{Repeated experience creates concepts}
Most machine learning algorithms work within a certain model or framework that allows certain parameters that are optimized to fit the data. We will argue that the mind on the other hand ``grows" circuits or functions bottom up -- this growth has no upper bound and can potentially continue endlessly creating more and more complex concepts. Thus by observing and processing external events, the mind can create computational frameworks that are not bounded by any specific model.

\noindent{\bf Relation to Kolmogorov Complexity (Finding succinct descriptions hierarchically):}
The same experience repeated many times is stored in memory as a hardened concept. Given a new episode of experience, we try to decompose it into concepts that are stored in memory. The principle of contrast is used to break up an experience into components that may be mapped to known concepts. Over time this produces a hierarchy of concepts build on top of each other. Thus the mind tries to come up with the simplest description of an experience by building concepts incrementally. This is similar in spirit to the notion of Kolmogorov complexity~\cite{Kol} that is the length of the turning machine with the shortest description that produces the string upon execution. The Kolmogorov complexity is incomputable as one cannot decide if a turing machine will terminate. Our mental process also attempts to find short descriptions of the input that can be easily obtained from previous concepts; thus it can be viewed as a heuristic for upper bounding the Kolmogorov complexity. Thus we are reaffirming the principle of simplicity that was proposed in ~\cite{Chater99}

For example, when we look at a room, we decompose the image into `known objects' and describe the scene as consisting of those objects in  a certain arrangement. Each object in-turn may be stored as a composition of prior concepts. Perhaps we store concepts as  programs, rather as circuits that recursively describe an input. Thus we may be able to think of mind as growing functional programs rather  functional circuits hierarchically as a heuristic for bounding the Kolmogorov Complexity of a stream of inputs. 

Kolmogorov Complexity is the ideal compression of an input. Thus the ideal efficient {\em approximation}-algorithm (indeed, a very gross approximation) for Kolmogorov Complexity can also be viewed as a univeral-compression algorithm. Since there is  a well known (see for example ~\cite{learning}) strong connection between compression and learning/prediction and communitation, it can also be viewed as a universal-learner for prediction. Thus the ideal cognitive process will simultaneously achieve good compression (that also facilitates quick communication) and good predictive power.
 
\noindent{\bf Lambda Calculus :} A concept can be viewed as some type of a lambda expression. Lambda calculus~\cite{lambda} is one of the oldest formalizations of functional languages with anonymous functions that can take other functions as arguments; that is it can build functions built on top of other functions. Each concept is obtained by composing previous concepts which can be thought of as composing the lambda expressions corresponding to those concepts.

\noindent{\bf Concepts of space, time and persistence of objects:}
Early on a child develops the concept of space and objects and the persistence of objects (that objects don't simply disappear)~\cite{Bloom}. The mind develops the ability to visualize objects and visualize the operations of moving objects, translating and rotating them in space -- thus given a scene it knows how it will change if you add another object somewhere or move a given object.

\noindent{\bf Parsing a simple Picture:}

Consider as an example how a child may process a simple image containing a rectangle colored blue and a triangle colored red partially overlapping. A large number of complex computations are involved in doing this.

\begin{enumerate}
\item To parse this picture a child needs to develop first the concept of rectangle and a triangle.
\item To understand a rectangle, it needs to understand the concept of a straight line segment.
\item When the child sees many instances of lines, it abstracts a common theme, that of moving in the same direction. This common theme present in the 'experience' of seeing a line, is stored as a concept of a 'straight line'
\item The concept of rectangle is built upon the concept of lines as using four lines arranged in a certain fashion -- two vertical and two horizontal lines properly aligned. Thus it makes use of concepts of vertical and horizontal that must have been learned previously.
\item Then the child must have experience with superimposing shapes on a blank paper. This is learnt by seeing many examples of objects being placed on a table.
\item Having known the concepts of rectangle and triangle, the child notices that the scene is obtained by first place a blue rectangle, and then a red triangle on top overlapping it partly. This then becomes a succinct description of the picture.
\end{enumerate}

A scene or an experience may be stored to different levels of detail depending on the level of attention. At first we get a crude description of the scene. At the next level of attention we refine the description to add more detail. This process may continue till we are able to completely describe the scene. If we do not revisit the experience for a long time, we may forget it, perhaps losing the fine details first and then going down the different levels of details in the description of the scene.
Thus the picture may be described in increasing detail starting from the most crude description to the fine one.
\begin{enumerate}
\item Some colorful objects on a white paper
\item A blue object and a red object
\item Objects are described by straight lines
\item The objects are a triangle and a rectangle
\item Some information about the position of the objects; such as, the rectange is in the middle and the triangle is to the upper left.
\end{enumerate}

The above sequence can be viewed as a circuit with 6 nodes, where each node applies a certain operation on the concept corresponding to the chain instructed so far. On first exposure one may construct the circuit 1. On next one may construct $1 \rightarrow 2$. Then $1\rightarrow 2\rightarrow 3$ and so on until we construct the fill circuit that completely describes the picture. Here the terms such as `rectangle', `triangle', `red', `blue', `object' are references to prior learnt concepts.

The memory thus stores a web of concepts connected by edges. Edges indicate the use of one concept in describing another concept

When two concepts co-occur frequently, it gives rise to a compound concept that references these two concepts. For example if `eating icecream' is frequently followed by `pleasure' it produces the concept `eating icecream is followed by pleasure' that references the two concepts. This new node connects the nodes `eating icecream' and `pleasure' in the web of concepts. Because of this the concept `eating icecream' becomes pleasurable. Thus nodes become pleasurable or painful to the extent to which they are near the concepts of `pleasure' or `pain' in the web of concepts graph.

\noindent{\bf An apple falling down :}

Consider an experience that involves watching an apple falling. This involves a sequence of events that produces several concepts in a sequence combined to produce the final concept of a ``falling apple". First contrast is used to identify a certain object of red color that is different from the rest of the color. Based on the shape it is mapped to the concept of an ``apple" stored in memory. Watching the image over time, one observes that the red part of the image is translating slowly in the down direction. This produces the concept of a ``persistent object" , the apple, moving down creating the concept of ``falling apple".

The arising of the percept of ``falling apple" may trigger additional percepts. If the person has had prior experience of eating an apple and finding it tasty, then the concept "apple is tasty" combined with the "falling apple" experience triggers the concept of "can reach out to eat a tasty apple". This would result in the recalling the emotion of "pleasant taste" from memory. Thus retrieval of a concept from memory tends to retrieve related concepts that are nearby in the graph as well.

Thus it seems there is a universal set of emotions, a universal set of sensations, a universal set of concepts, and a universal set of rules for the arising of concepts from sensations, and the arising of emotions from concepts, depending on the contents in the memory. The sequence of receiving sensations, arising of concepts, and emotions, in turn triggering additional concepts and emotions in a chain becomes an {\em episode of experience}. The sets of sensations, concepts, and emotions, may include simple elemental ones such as sensation of ``red color", ``sweet taste", ``pain", ``pleasure", ``object", ``space" and compound ones such as ``a given colorful scenery", ``combination of tastes", complex emotions, and realization of complex phenomena.

\section{Concepts grow like circuits hierarchically}
Exposure to external events create sensations that result in an experience. Alternately an experience may also be created by recalling events (experiences) from memory.
The final experience depends on how the sensations are interpreted based on the web of concepts stored in memory.

An experience is broken down into smaller units of experience based on contrast. It is then interpreted as a composition of a small number of prior concepts. A common experience or a common theme in prior experiences is instantiated as a new concept. A frequently visited concept is hardened and ingrained in memory. The circuits or functions underlying these frequent concepts can be computed very quickly as if they were in the `fast path'. Thus concepts move from slow path to fast path based on usage. Concepts in the slow path need to evaluated consciously and slowly. Concepts in the fast path are evaluated 'automatically' without a conscious experience of the computation.

The concept-graph operations can be described thus:
\begin{verbatim}
For each new experience from external events:
    Break the experience into components based on contrast (recursively).
    Check if the new experience can be described as a simple composition of a small 
    number of prior concepts in the concept-graph
    If so, create a new concept (node) as a circuit that composes these prior concepts
\end{verbatim} 

Thus we can {\em approximately} view concepts as some type of functions or (lambda expressions) $f_1, f_2,...,f_n$. At any time if a new repeated experience $e$ can be approximated as a composition of a small number of some of these functions $e \approx f_i . f_j .f_k$ then we create a new node $f_i . f_j .f_k$ in the hierarchy. Any future exposure to experience $e$ will trigger this new node. The more $e$ is experienced the more weight this node accumulates (it perhaps also decays geometrically with disuse.) Observe that if the ``." operator were simply the concatenation operator instead of a composition we would only get a context-free-grammar. But since we are allowing a composition we can get arbitrary turing machines.

When visiting candidate concept to explain the new experience, we visit them in order of their weight. High weight concepts are tried first. The weight of a concept depends on how often it has been invoked in the past or how closely related it is to recently visited concepts. The rank or order in which functions are tried may also depend on the outputs of the previously tried functions (that provide a context). The choice of the first function $f_i$ may condition the choice of the next function for composition based on how compatible they are. The ranking of functions is a crucial component in the efficiency (and efficacy) of learning new concepts.

As a concept's get matched, they are fired as output that may then become inputs for more advanced concepts built on them. The process can be viewed as many base level concepts feeding into a circuit and outputting a complex concept at the top of the circuit.

Thus on hearing a sequence of sounds, contrast is used to break it into pieces of continuous sounds separated by silent pauses. Each continuous stream of sound is broken into syllables by contrasting the nature of the sound. Similarly a picture is broken into components by contrasting the colors or a sharp change in intensity of color. Each component (syllable or image) may be mapped to concept. Group of concepts are mapped to a compound concept and this continuous recursively. Attention of the mind is attracted whenever an experience is parsed into a concept. The more complex the experience and simpler the concept maps to the higher the level of attention that it attracts~\cite{Chater99, CV03}; Simplicity theory, points out several examples where this is evident. If many concepts (circuits) have a common component, then the mind may observe this commonality and abstract the common component. This common component is used to simplify all the previous circuits.

{\bf Storing concepts in increasing levels of detail:}
At first exposure, an experience may only be partially described in terms of prior concepts. On subsequent exposures, this description is refined giving more detailed descriptions. This continues until we may get a perfect description of the experience in terms of prior concepts. Thus we obtain a chain of refinements each making a small change on the prior concept where the first node in the chain is a crude first impression of the experience and the last one perfectly matches the experience.

{\bf How associations are created :}
If two concepts occur frequently in succession, then the experience of ``B follows A" becomes a common experience and thus a new concept. Many such experiences of ``x follows y" has the common theme of ``some concept follows another concept". Thus ``follows" becomes a new node that generalizes all ``B follows A" types of nodes. That is ``B follows A" is viewed as ``follows" experience applied on the previous concepts B and A.
There are different concepts (functions) such as ``follows in time", ``follows in space", ``near by in time or space" describing different types of associations.

\noindent{\bf Learning numbers :}

Here is how the concept of the number ``2" may be leant incrementally.
Consider a picture containing 2 circles another with 2 squares and another with 2 apples. The concept of the number ``2" is distilled as a common experience in all these pictures. Before learning the number, the pictures may be described as follows.

draw a circle, repeat, 

draw a square, repeat.

draw an apple, repeat.

The common description of operations ``object, repeat" becomes a concept that we associate with the symbol ``2" and the sound of the word ``two".
Thus the concept ``2" is (lambda) function that given an object produces two copies of that object $ (\lambda x ) x x$.

A child gets more pleasure from eating two candies than eating one candy, thus it contrasts the image of a plate containing two candies from that containing one candy.
Similarly it contrasts three candies from two candies and understands the concept that `a configuration of 3 candies' is 'better than' a 'configuration of two candies. This produces the concept of ``greater than".'
Similarly the abstract concept of each of the different numbers `3', `4', `5' is produced. A canonical image containing `some quantity' of candy may be used to represent an arbitrary number not known precisely.

\noindent{\bf Learning Addition :}

The concept of `add one candy' is learnt by observing constantly observing the action of `taking another candy and putting it on the plate', and noticing that by repeating the `add one candy' action one obtains different configurations of candy on the plate.

Similarly the concept of `add one apple' may be learnt. And then one may observe the commonality of these operations and create a new abstract concept of `add one object' that given a scene containing a collection of objects adds
another copy of the object to the scene. The operation may make use of some canonical object to represent any object.

Similarly the concepts of subtraction, multiplication and division are built incrementally on top of each other. This is similar to how one could write a function defining these operations in a modular fashion from the simplest to the most complicated; except that the creating of the functions happen by observing examples and exacting out the simple new operation on existing concepts that describes these examples ~\cite{example}.

One develops not only the ability to manipulate spatial configurations but also imaginary configurations of abstract concepts. For example one may manipulate numbers once the concept of numbers and concept of operations on these numbers is developed. Such a sequence of mental manipulations is also an episode of experience that can result in new concepts. 

For example by thinking about numbers, multiplication and division, one may come up with the concept of prime numbers. Upon further contemplation or through further learning one may realize that there is endless prime numbers. Each such realization is an experience that results in a new concept. It is used to describe the "scene" involving the manipulation of numbers. By looking at things moving in a straight line, we produce the concept of an ideal straight line even though the lines we saw were never perfect but were approximated by the `abstract line' which is a simple description for these real lines. By looking at longer and longer lines, we generate the notion of `infinite line' as this captures the experience of a very long lines which seem to be similar to an `infinite line' -- for example, travelling on a very long road feels just like an endless road. Thus we learn the concept of `infinity'.

Exposure to a sequence of experiences results in the growth of concepts. This can be viewed as the growth of circuits (or algorithms) that are used to describe (or make sense) these experiences. Over time one gets a complex web of circuits interconnected as a graph of concepts.

\noindent{\bf Concept of ``War"}

``War" is a complex concept build upon other complex concepts. To understand this one first has to develop the concept of ``two people fighting" -- this in turn happens through experience of two people with conflicting wishes subject to the emotion of anger that leads to explicit acts of hurting one another. Then one has to develop the notion of ``Nations" and the concept of ``wishes and aspirations of nations". Then one builds the concept of ``Nations fighting". One then sees certain typical images of a battle and associates this image with the concept of ``Nations fighting". This results in the compound concept of ``War"

\noindent{\bf Concept of ``Contentment"}

This is another complex concept that develops hierarchically over time. First a child needs to understand the concept of ``possessions" -- a common pattern in the experiences of ``my toys", ``my books". Then one sees the arising of delight from new possessions and then the fading away of the novelty and delight over time. This gives rise to the concept of ``dissatisfaction". When one experiences a repeated pattern of ``new possessions provide delight and then become dissatisfactory" one develops the common theme of ``discontent". This makes one appreciate the concept of ``contentment".

\noindent{\bf Communication is about recreating circuits}

Communication is about transferring mind states. Its about recreating an experience from the speaker to the listener. This is done by creating the final concept hierarchically in the listener. If the listener lacks many of the underlying layers of concepts then these need to be created bottom up. This is how we teach children -- we communicate concepts gradually one by one starting from ground zero.

\subsection{Emotions}
Each experience may be associated with one or more emotions. There are specific circuits output emotions based on the type of experience. For example:
When an experience matches the template ``Another persons actions led to pain for me" it triggers the emotion of ``anger".

Observe that here ``person", ``action", ``led", ``pain", ``me" are all prior existing concepts. The above template thus identifies a certain sub-structure in the circuit representing the experience.

``tried hard to get a certain pleasure but failed repeatedly" triggers "frustration".

Similarly, some circuit perhaps maps experiences to ``funny".
There are certain primitive functions that map the space of emotions to pleasant or painful feelings of different magnitudes. Concepts are described as pleasant or painful depending on their association with the two primitive concepts of ``pain" and ``pleasure". This could be based on association or based on proximity in the concept graph.

Some experiences may be described as ``smooth or soft". For example the touch of petals, or more metaphorically soft music, or a smoothly-flowing story. A certain circuit distills this common feature from these apparently diverse experiences. Such a circuit may simply be looking for the absence of a discontinuity in the derivative in the intensity of the underlying feelings as a function of time. For example, touch is smooth if the surface doesn't have sharp bends or if the resistance felt on applied pressure doesn't have a discontinuous derivative. A soft story may be one where the stream of emotions produced there is no wild or sharp changes.

Emotions are a key component in the formation of a concept. For example the concept of ``getting fired from a job" is more than just termination of employment. The experience of the feeling of rejection and the fear of financial instability are really the most important part of this concept. It is often the emotional component that attracts the minds attention to the concepts. Concepts without a strong emotional content will fail to grab attention and thus will not be repeated enough to be ingrained in memory.

Sometimes, instead of building a new circuit, we simply uncover a circuit that is implicit in the production of a certain emotion. For example there is pre-existing circuit that governs the production of the emotion ``anger". The circuit implicitly encodes the concepts of ``hurt" and ``another persons actions". As we experience anger, we observe the patterns we explicitly are able to construct these concepts and their corresponding circuits. Thus this construction can be viewed as not really constructing a new circuit, but uncovering the black box circuit implicit in the production of ``anger".

These circuits process not only discrete concepts but also continuous ones. Consider for example how one would teach a child about real numbers. A child is told to observe a ball rolling a notice that it does not move in steps but continuously where the distance travelled ``smoothly" increases over time. This simple observation provides a rich intuition for real numbers which forms the basis for other advanced mathematical concepts such as co-ordinate geometry, vectors and calculus. These tools are used not only in situations involving continuous concepts but also discrete ones: for example to understand solutions to the discrete problems of TSP or Graph-Coloring one may make use of linear programming, a method that uses the concept of real numbers.

The circuits process concepts that cannot simply be described as discrete or continuous phenomena. Consider for example the reaction of a person when she first sees turbulent waves at the sea shore. While some features may be extracted to find similarities to previous concepts, perhaps much of the experience is stored as ``raw experiential data". This new concept may later be used to describe other seemingly unrelated experiences. For example one may experience some ``turbulent emotions" and may see the similarity between the ``turbulent state of mind" and the ``turbulent waves in the ocean".

Consider a tumbler containing a mixture of various chemicals subject to heat and light. The way the chemicals react under different amounts of heat and light may be modeled using a circuit (perhaps analog circuit) that captures the rules according to which the chemicals react to produce new ones. Thus a chemical reaction may be represented using a computational node that takes the reactants as inputs and outputs the resultant compounds. The circuits in the mind are certainly at at least as complex as the ones needed to model chemical reactions.

Sometimes complex pathways of circuits can be formed by a sequence of instinctive emotions lined up that cause an individual to behave in a certain way. Consider for example the computations involved in a baby trying to suckle for milk. The circuits for searching for the nipple, inserting into the mouth, sucking on it need to be formed or ``uncovered". Even though the child may not be initially aware of the full circuit of concepts leading to the satisfaction of milk, at each step it feels an instinctive urge to perform each step feeling some kind of a gradual reward culminating in the final big reward of food that results in a pleasant feeling. This reinforces the final circuit that the child needs to suckle for milk when it is hungry. Thus the carefully aligned sequence of instincts, the different feelings, the warmth of the skin, the feeling from latching on to the nipple, suckling, and the satisfaction gushing of milk are lined up in perfect order to create the complex circuit of concepts needed to derive milk.

\section{Experiences as learning an ensemble of functions}
Concept formation can be compared to learning an ensemble of functions. Say there are many functions -- and we are given certain input, output pairs for each of these functions. Our goal is to learn these functions. To distinguish these functions we will say these functions have different colors -- say the red function maps input (1,3) to output 4 input (2,3) to 5, the green function maps input (2,4) to 8 (3,4) to 12. These functions could be interpreted as experiences in different contexts. The input output pairs of the different functions may arrive interleaved in arbitrary order. An individual function may be complex and thus one may not be able to learn the function from its examples. But if these functions form a hierarchy where function can be expressed as a simple composition of a few lower functions in the hierarchy then over time, a simple algorithm can be used to learn all these functions. First one can explain the leaf level functions. Having learn't these concepts, one can explain the functions at level 2 and so on and so forth till one has learnt all the functions. Note that if one is given only instances of one function in isolation, then it would be much harder to learn the function.

Similarly, as we interact with the world, it presents us several functions in different in different contexts that we visit. First we learn the easiest concepts and then build the higher level concepts. There is a pre-existing algebra over the universe of concepts that allows one to compose a number of concepts resulting in another compound concept. This algebra is used to construct the concept hierarchy from experience. The external world provides us with several instantiations of functions built upon each other hierarchically. For example the sight of  ``leaves” are a common pattern that we recognize. ``A collection of leaves on top of a trunk” is the next higher level pattern that we identify as a tree. We notice flowers and fruits on trees. We notice seeds in the fruits and the seeds falling in the ground and giving rise to small plants that grow into big trees. Thus the final high level complex pattern is built upon several lower level patterns. Of course, this continues endlessly into the learning more advanced concepts. For example, we notice that the notion of the seed reproducing a tree is similar to an egg producing an animal which informs us about the concept of ``reproduction of living things”. Perhaps all of our thought process can be understood in this way.

\subsection*{Acknowledgements}
We would like to thank Alexander Andoni, Ollie Williams and Aviv Zohar for insightful discussion.

\end{document}